\documentclass[letterpaper]{article} 
\usepackage[preprint]{aaai2027} 
\usepackage[hyphens]{url} 
\usepackage{graphicx} 
\usepackage{natbib} 
\usepackage{caption} 
\usepackage{amsmath}
\usepackage{amsthm}
\usepackage{amssymb}
\usepackage{booktabs}
\usepackage{algorithm}
\usepackage{algorithmic}
\usepackage{subcaption}

\title{Phase-Consistent Magnetic Spectral Learning for Multi-View Clustering}

\author{
Mingdong Lu,
Zhikui Chen,
Meng Liu,
Shubin Ma,
Zhengyang Tang
}

\affiliations{
School of Software, Dalian University of Technology\\
\{mingdonglu, liumeng, shubinma, zhengyangtang\}@mail.dlut.edu.cn,
zkchen@dlut.edu.cn
}

\begin{document}

\maketitle
\begin{abstract}
Unsupervised multi-view clustering (MVC) aims to partition data into meaningful groups by leveraging complementary information from multiple views without labels, yet a central challenge is to obtain a reliable shared structural signal to guide representation learning and cross-view alignment under view discrepancy and noise. Existing approaches often rely on magnitude-only affinities or early pseudo targets, which can be unstable when different views induce relations with comparable strengths but contradictory directional tendencies, thereby distorting the global spectral geometry and degrading clustering. In this paper, we propose \emph{Phase-Consistent Magnetic Spectral Learning} for MVC: we derive a confidence-directed cross-view flow from anchor assignments, encode it as a phase term, and combine it with a nonnegative magnitude backbone to form a complex-valued magnetic affinity, extract a stable shared spectral signal via a Hermitian magnetic Laplacian, and use it as structured self-supervision to guide unsupervised multi-view representation learning and clustering. To obtain robust inputs for spectral extraction at scale, we construct a compact shared structure with anchor-based high-order consensus modeling and apply a lightweight refinement to suppress noisy or inconsistent relations. Experiments on ten public benchmarks show that our method achieves the best result on 22 of 30 dataset--metric combinations and ranks among the top two on 27 combinations, demonstrating strong and broadly consistent performance against classical and recent MVC baselines.
\end{abstract}

\section{Introduction}

Unsupervised multi-view clustering aims to partition data into semantically meaningful clusters by exploiting complementary information from multiple views, without requiring manual annotations.
Such multi-view observations widely arise in practice, e.g., multi-sensor measurements or heterogeneous feature extractors of the same data.
While multi-view learning can improve robustness via cross-view complementarity, achieving \emph{consistent} and \emph{robust} clustering in the unsupervised setting remains challenging.

A common thread behind existing MVC approaches is to infer a \emph{shared clustering structure} (e.g., a shared embedding/graph or pseudo targets) across views and use it to guide representation learning and fusion~\cite{ZhouDLWD24,FangLLGJZ23}.
However, in real-world multi-view data, noise, view discrepancy, and structural conflicts can make such shared signals unreliable.
When pseudo targets or structural cues are inaccurate, directly enforcing strong cross-view alignment may amplify errors and cause supervision drift or mismatched alignment~\cite{Xu_2024_CVPR,arazo2020pseudolabeling,gao2024semanticinvariance}.
Therefore, constructing a reliable shared structural signal under view conflicts and turning it into stable learning supervision remains a key bottleneck for unsupervised MVC.

\begin{figure}[t]
    \centering
    \begin{subfigure}[t]{0.49\linewidth}
        \centering
        \includegraphics[width=\linewidth,trim=50mm 10mm 50mm 10mm,clip]{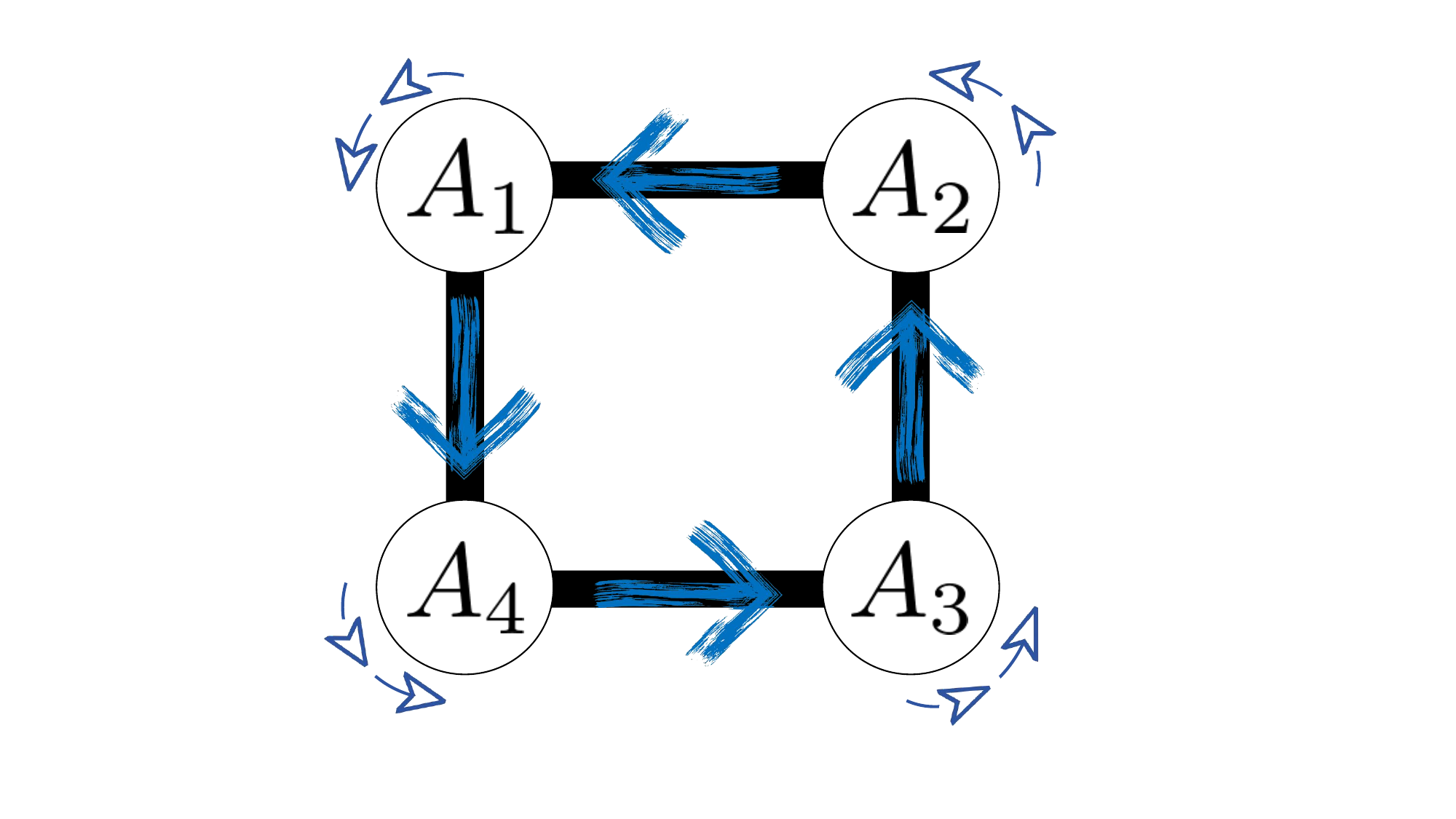}
        \caption{Consistent phase}
        \label{fig:phase-toy-a}
    \end{subfigure}\hfill
    \begin{subfigure}[t]{0.49\linewidth}
        \centering
        \includegraphics[width=\linewidth,trim=50mm 10mm 50mm 10mm,clip]{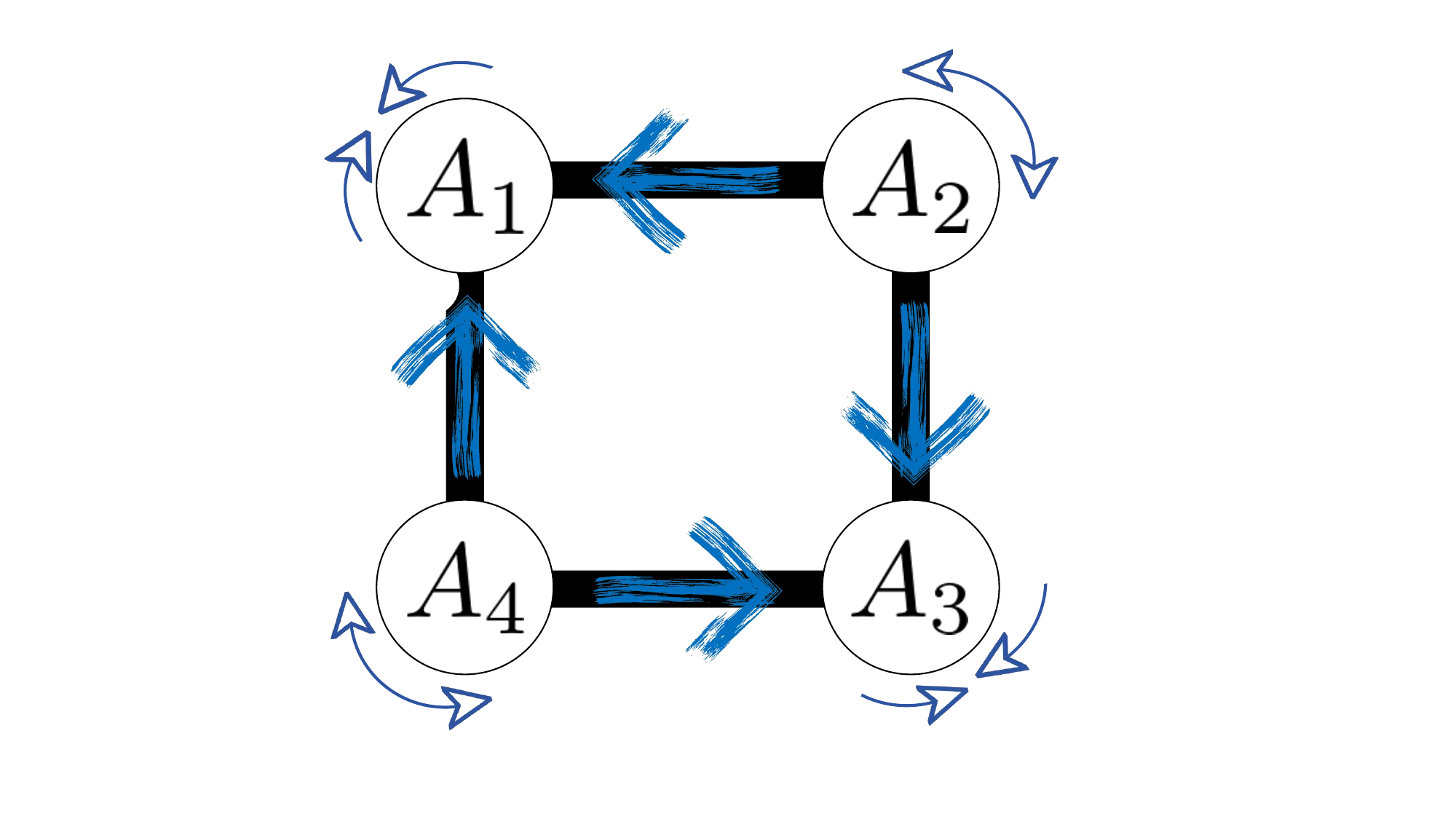}
        \caption{Conflicting phase}
        \label{fig:phase-toy-b}
    \end{subfigure}
    \caption{\textbf{Phase effects beyond affinity magnitude.}
    With identical affinity magnitudes, consistent directions yield a coherent flow and a stable global structure, whereas conflicting directions cancel out and induce a different spectrum.}
    \label{fig:phase-toy}
\end{figure}

More importantly, instability does not necessarily manifest as weak connections.
Beyond connection strength (magnitude), cross-view relations may exhibit \emph{directional tendencies} whose agreement can substantially affect global spectral behavior.
Here the ``direction'' is not intrinsic to the original undirected affinity; it is induced by the relative confidence of view-wise anchor assignments and can be encoded as a phase term.
This is consistent with directed-graph and connection-Laplacian spectral theories~\cite{chung2005directed,singer2012vdm}, and magnetic (complex-valued) Laplacians that incorporate phase/direction to characterize directed-network structures~\cite{fanuel2017magnetic}.
As illustrated in Fig.~\ref{fig:phase-toy}, even under identical affinity magnitudes, consistent directions yield a coherent flow and a stable spectrum, whereas conflicting directions can cancel out and induce a markedly different spectrum.
This motivates us to model not only magnitude but also confidence-directed cross-view flow when constructing shared structural signals for MVC.

To address these challenges, we propose \emph{Phase-Consistent Magnetic Spectral Learning} for unsupervised MVC.
We construct a nonnegative magnitude backbone as a stable shared geometry, encode confidence-directed cross-view flow as a phase term to form a magnetic affinity, and extract a shared spectral signal via a Hermitian magnetic Laplacian to provide structured self-supervision for multi-view representation learning.
For scalable and robust structure estimation, we employ an anchor-based high-order consensus model (anchor hypergraph) and an efficient refinement to suppress noisy or inconsistent relations before phase estimation and spectral extraction.

Our main contributions are summarized as follows:
\begin{itemize}
    \item We propose \emph{phase-consistent magnetic spectral learning} for unsupervised MVC by jointly modeling magnitude and confidence-directed cross-view flow, and deriving a stable shared spectral signal via a Hermitian magnetic Laplacian for structured self-supervision.
    \item We develop an \emph{anchor-hypergraph} based structure construction that preserves high-order cross-view consensus while enabling spectral learning in a compact anchor domain, together with an efficient refinement for robust phase estimation and spectral extraction.
    \item Experiments on ten benchmarks demonstrate leading or highly competitive
performance against classical and recent MVC baselines, while extensive
analyses confirm the effectiveness of the proposed phase-aware spectral
modeling and its key components.
\end{itemize}

\section{Related Work}

\paragraph{Multi-view clustering and scalable structure learning.}
Unsupervised multi-view clustering seeks a shared partition from
complementary views. Classical methods encourage consensus through
co-regularized spectral or subspace learning~\cite{kumar2011coreg},
whereas recent deep approaches learn view-specific representations
through self-training or contrastive objectives
\cite{xie2016dec,xu2022mflvc,Cui2024DCMVC,Chen_2023_ICCV}.
Graph-based methods further construct shared bipartite or anchor
graphs to capture cluster geometry efficiently
\cite{li2015bipartite,dmcag23,zhang2024camvc,wang2025dmac}.
Recent studies further mine cross-view complementarity and
consistency, disentangle shared and view-specific factors, or address
incomplete and unaligned views through semantic label diffusion
\cite{liu2026idccm,li2026dmvcs,CHEN2026123626}; BONE bridges
optimization and neural networks for efficient MVC
\cite{xu2026bone}.
Anchor and hypergraph models provide scalable representations and
capture high-order consensus
\cite{chen2010landmark,zhou2006hypergraph,feng2019hnn,zeng2026hrfal}.
However, these methods primarily rely on real-valued affinity
magnitudes and do not explicitly encode confidence-directed
cross-view discrepancies as a spectral phase.

\paragraph{Geometry refinement and phase-aware spectral learning.}
Curvature- and Ricci-based reweighting can suppress unreliable
connections and stabilize graph geometry
\cite{ollivier2009ricci,sia2019ricciflow,ni2018orflow}.
Meanwhile, directed and connection Laplacians demonstrate that
orientation information can fundamentally affect spectral
representations~\cite{chung2005directed,singer2012vdm}.
Magnetic Laplacians encode such information as complex-valued phases
while preserving Hermitian spectral properties
\cite{fanuel2017magnetic,resende2020magnetic,
fiorini2022sigmanet,hirn2021magnet}.
Unlike prior applications mainly designed for directed networks,
we derive a confidence-directed phase from multi-view anchor
assignments and use the resulting magnetic spectrum as structured
self-supervision for MVC.

\begin{table}[t]
\centering
\small
\setlength{\tabcolsep}{5pt}
\begin{tabular}{ll}
\toprule
Symbol & Meaning \\
\midrule
$V,\,n,\,K$
  & \#views, \#samples, \#clusters \\
$X^{(v)},\,Z^{(v)}$
  & input data and latent codes of view $v$ \\
$d$
  & latent dimension \\
$m_a,\,m$
  & anchors per view; total anchors $m=Vm_a$ \\
$A^{(v)}$
  & latent anchor centers of view $v$ \\
$C^{(v)}$
  & sample--anchor coefficients of view $v$ \\
$H$
  & anchor hypergraph incidence matrix ($m\times n$) \\
$w,\,\kappa$
  & hyperedge weights; curvature reweighting \\
$S'$
  & curvature-refined anchor affinity (anchor graph) \\
$\tilde{A},\,L_{\mathrm{mag}}$
  & magnetic adjacency and Laplacian on anchors \\
$U$
  & sample-level spectral embedding \\
$Q^{(v)},\,P$
  & per-view soft assignments; shared targets \\
\bottomrule
\end{tabular}
\caption{Notation for the main geometric and spectral components.}
\label{tab:notation}
\end{table}

\begin{figure*}[t]
    \centering
    \includegraphics[width=\textwidth,trim=10mm 50mm 6mm 50mm,clip]{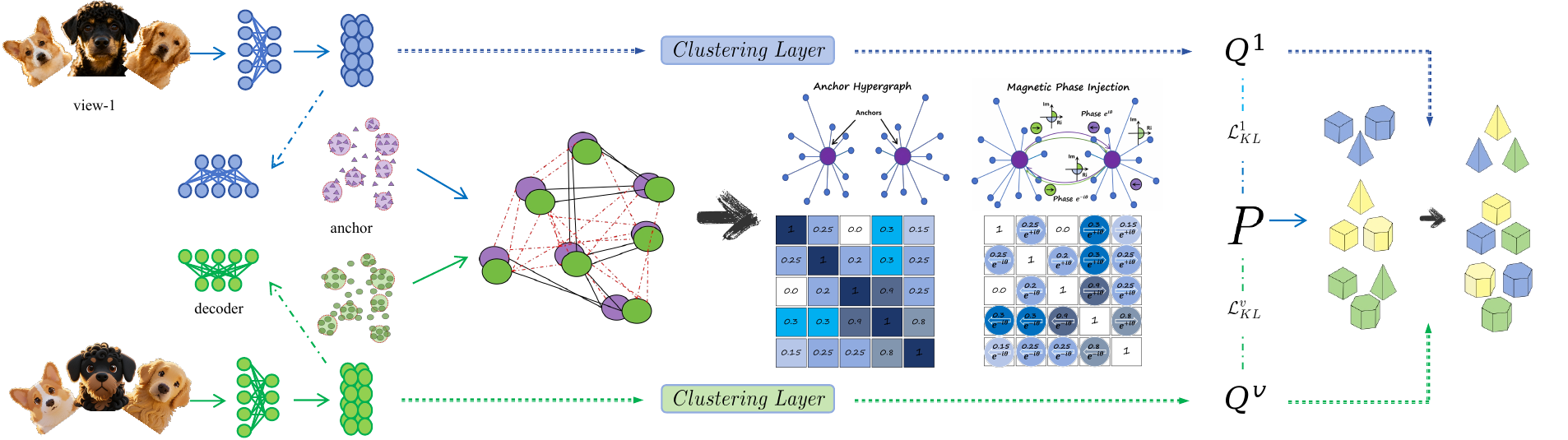}
    \caption{\textbf{Overall framework of Phase-Consistent Magnetic Spectral Learning for MVC.}
    Each view is encoded into latent features with reconstruction learning.
    View-wise latent anchors induce sparse sample--anchor relations, which are aggregated into a multi-view anchor hypergraph to form a compact shared magnitude backbone.
    We stabilize the backbone by refining hyperedge weights via curvature-inspired flow reweighting, then encode confidence-directed cross-view flow as a magnetic (phase) term and extract a shared spectral signal using a Hermitian magnetic Laplacian.
    The resulting signal defines a global target distribution $P$ to supervise per-view predictions $Q^{(v)}$ via KL alignment, while label contrastive consistency further reduces cross-view mismatch for robust clustering.}
    \label{fig:framework}
\end{figure*}

\section{Method}

\paragraph{Problem statement.}
We study unsupervised multi-view clustering.
Given $V$ views of the same $n$ samples, $\{X^{(v)}\}_{v=1}^{V}$, the goal is to partition the samples into $K$ clusters by learning view-specific latent representations $\{Z^{(v)}\}$ and, crucially, constructing a reliable \emph{shared structural signal} across views to guide representation learning and cross-view alignment.
The key symbols used in this section are summarized in Table~\ref{tab:notation}.

\paragraph{Method overview.}
Motivated by the instability of shared structure under view discrepancy and noise, we build a compact shared geometric backbone in the latent space and then progressively turn it into stable training signals.
Specifically, we first learn view-wise latent codes and construct an anchor hypergraph as a scalable magnitude backbone, which will be refined and later augmented with phase information for magnetic spectral signal extraction as illustrated in Figure~\ref{fig:framework}.

\subsection{Multi-view Autoencoders and Anchor Hypergraph Construction}
\label{subsec:anchor-hypergraph}


\paragraph{Multi-view autoencoders.}
For each view $v\in\{1,\dots,V\}$, we learn an encoder--decoder pair $(f^{(v)}_{\phi},g^{(v)}_{\psi})$ to obtain latent codes and reconstructions:
\begin{equation}
z_i^{(v)}=f^{(v)}_{\phi}\!\big(x_i^{(v)}\big),\qquad
\hat{x}_i^{(v)}=g^{(v)}_{\psi}\!\big(z_i^{(v)}\big),
\label{eq:ae-single-view}
\end{equation}
where $x_i^{(v)}\in\mathbb{R}^{d_v}$ and $z_i^{(v)}\in\mathbb{R}^{d}$.
We pretrain the autoencoders with the standard reconstruction objective
\begin{equation}
\mathcal{L}_{\mathrm{rec}}
=\sum_{v=1}^{V}\big\|X^{(v)}-\hat{X}^{(v)}\big\|_{F}^{2}.
\label{eq:rec-loss}
\end{equation}

\paragraph{Anchor-based subspace representation.}
To obtain a compact latent geometry, we summarize each view by the same number $m_a$ of latent anchors $A^{(v)}\in\mathbb{R}^{d\times m_a}$, initialized by running $k$-means on $\{z_i^{(v)}\}_{i=1}^{n}$.
Each latent code is approximated by a convex combination of anchors, $z_i^{(v)}\approx A^{(v)}c_i^{(v)}$, where $c_i^{(v)}\in\mathbb{R}^{m_a}$ is obtained by a simplex-constrained quadratic program:
\begin{equation}
\min_{c_i^{(v)}\ge 0,\;\mathbf{1}^\top c_i^{(v)}=1}
\ \big\|z_i^{(v)}-A^{(v)}c_i^{(v)}\big\|_2^2+\gamma\big\|c_i^{(v)}\big\|_2^2.
\label{eq:anchor-qp}
\end{equation}
Stacking $\{c_i^{(v)}\}_{i=1}^n$ as columns forms $C^{(v)}\in\mathbb{R}^{m_a\times n}$.

\paragraph{Anchor hypergraph construction.}
We use the same $m_a$ in every view, so $m=Vm_a$ and the view-wise assignment confidences are directly comparable. We independently retain the $r$ largest entries in each view-specific coefficient column and then concatenate the sparse blocks to obtain $H\in\mathbb{R}^{m\times n}$:
\begin{equation}
H=\mathrm{Concat}_{v=1}^{V}\!\big(\mathcal{T}_r(C^{(v)})\big),
\label{eq:incidence}
\end{equation}
where $\mathcal{T}_r(\cdot)$ denotes column-wise Top-$r$ sparsification. Thus, each sample contributes at most $Vr$ nonzero incidences, and each column $H(:,i)$ defines a sample-associated hyperedge. We initialize hyperedge weights with $w^{(0)}=\mathbf{1}\in\mathbb{R}^{n}$; the resulting $(H,w^{(0)})$ will be refined by curvature-inspired reweighting and used as the shared geometric backbone in the sequel.

\subsection{Phase-Consistent Magnetic Spectral Signal and Self-supervision}
\label{subsec:curvature-spectral}


\paragraph{Curvature-based refinement of the anchor hypergraph.}
Given Eq.~\eqref{eq:incidence}, let $d_a=\max\{\sum_eH_{ae},1\}$ and $\delta_e=\max\{\sum_aH_{ae},1\}$. We define the incidence-based score
$\kappa_e=\delta_e^{-1}\sum_aH_{ae}d_a^{-1/2}-m^{-1}\sum_a d_a^{-1/2}$ and update
\begin{equation}
\begin{aligned}
\hat w_e^{(t+1)}&=\max\{w_e^{(t)}(1-\tau_{\mathrm{RF}}\kappa_e),\epsilon\},\\
w_e^{(t+1)}&=n\hat w_e^{(t+1)}/\textstyle\sum_{e'}\hat w_{e'}^{(t+1)}.
\end{aligned}
\label{eq:ricci-update-short}
\end{equation}
The second line preserves the total weight. After $T_{\mathrm{RF}}$ iterations we obtain $w^{(T_{\mathrm{RF}})}$.

\paragraph{Curvature-refined anchor affinity.}

Let $\delta=H^\top\mathbf{1}_m$ and $\delta_\epsilon=\max(\delta,\epsilon\mathbf 1)$, and define $D_e=\mathrm{diag}(w^{(T_{\mathrm{RF}})}\oslash\delta_\epsilon)$. We first form $\bar S=HD_eH^\top=\tilde C^\top\tilde C$ with $\tilde C=D_e^{1/2}H^\top$, and remove self-connections:
\begin{equation}
S'=\bar S-\mathrm{diag}\!\big(\mathrm{diag}(\bar S)\big)\in\mathbb{R}^{m\times m}.
\label{eq:anchor-affinity-short}
\end{equation}
This factorization enables spectral computation in the compact anchor domain.

\paragraph{Phase-consistent magnetic spectral embedding.}
To encode cross-view directional tendencies, we augment the refined magnitude backbone $S'$ with a magnetic phase.
For each sample $s$ and view $v$, let
\begin{equation}
a_s^{(v)}=\arg\max_{a} C^{(v)}_{a,s},
\qquad
\rho_s^{(v)}=\max_{a} C^{(v)}_{a,s},
\label{eq:anchor-confidence}
\end{equation}
denote the top-assigned anchor and its confidence. Its global index is
$b_s^{(v)}=a_s^{(v)}+(v-1)m_a$. For each unordered view pair, let
$\Delta\rho_s^{(v_1,v_2)}=\rho_s^{(v_1)}-\rho_s^{(v_2)}$;
the confidence difference determines the flow direction:
\begin{align}
\tilde{F}_{ij}
={}&
\sum_{s=1}^{n}
\sum_{1\leq v_1<v_2\leq V}
\Delta\rho_s^{(v_1,v_2)}
\Big(
\mathbb{I}\!\left[i=b_s^{(v_1)},\,j=b_s^{(v_2)}\right]
\nonumber\\[-1mm]
&\hspace{31mm}
-\mathbb{I}\!\left[i=b_s^{(v_2)},\,j=b_s^{(v_1)}\right]
\Big),
\label{eq:flow-matrix}
\end{align}
which directly satisfies $\tilde{F}^\top=-\tilde{F}$.
We then rescale it to a phase matrix:
\begin{equation}
\Theta=\beta\frac{\tilde{F}}{\max_{i,j}|\tilde{F}_{ij}|+\epsilon},\qquad
\beta=\pi q,\; q\in[0,1].
\label{eq:phase-matrix-short}
\end{equation}
Finally, we restrict phases to existing edges of $S'$:
\begin{equation}
\Theta\leftarrow \Theta\odot \mathbb{I}[S'>0],\qquad
\Theta\leftarrow \tfrac12(\Theta-\Theta^\top).
\label{eq:phase-mask}
\end{equation}
The complex-valued magnetic adjacency is
\begin{equation}
\tilde{A}=S'\odot \exp\!\big(\mathrm{i}\,\Theta\big),
\label{eq:mag-adjacency-short}
\end{equation}
Since $S'^{\top}=S'$ and $\Theta^{\top}=-\Theta$, the magnetic adjacency satisfies $\tilde{A}^{\mathrm{H}}=\tilde{A}$.
With degree matrix $D_{\tilde{A}}=\mathrm{diag}(|\tilde{A}|\mathbf{1})+\epsilon I$, the Hermitian magnetic Laplacian is
\begin{equation}
L_{\mathrm{mag}}=I-D_{\tilde{A}}^{-1/2}\tilde{A}D_{\tilde{A}}^{-1/2}.
\label{eq:lmag-short}
\end{equation}
We take the $K$ eigenvectors of $L_{\mathrm{mag}}$ with smallest eigenvalues to form an anchor embedding
$\Phi_{\mathrm{mag}}\in\mathbb{C}^{m\times K}$ and lift it to samples via
\begin{equation}
M=\mathrm{diag}(\delta_\epsilon)^{-1}H^\top,\qquad
U_c=M\Phi_{\mathrm{mag}}.
\label{eq:lift-to-samples}
\end{equation}
Finally, we output a real embedding by concatenating real/imaginary parts and normalizing rows:
\begin{equation}
U=\mathrm{row}\text{-}\mathrm{norm}\big([\Re(U_c),\,\Im(U_c)]\big)\in\mathbb{R}^{n\times 2K}.
\label{eq:sample-embed-short}
\end{equation}

\paragraph{Pseudo-labels and spectral self-supervision.}
Given $U$, we initialize cluster centers $\{\mu_j\}_{j=1}^{K}$ by $k$-means and compute a Student-$t$
soft assignment $Q=[q_{ij}]\in\mathbb{R}^{n\times K}$:
\begin{equation}
q_{ij} =
\frac{\big(1 + \|u_i - \mu_j\|_2^2 / \alpha\big)^{-(\alpha+1)/2}}
{\sum_{j'} \big(1 + \|u_i - \mu_{j'}\|_2^2 / \alpha\big)^{-(\alpha+1)/2}}.
\label{eq:dec-q-short}
\end{equation}
We then form a sharpened target distribution $P=[p_{ij}]$ to emphasize confident assignments:
\begin{equation}
p_{ij} =
\frac{ q_{ij}^2 / \sum_{i'} q_{i'j} }
{\sum_{j'} q_{ij'}^2 / \sum_{i'} q_{i'j'} }.
\label{eq:dec-p-short}
\end{equation}
Each view applies the same Student-$t$ head to $z_i^{(v)}$ and trainable centers $\{\nu_j^{(v)}\}$, i.e., $q_{ij}^{(v)}\propto(1+\|z_i^{(v)}-\nu_j^{(v)}\|_2^2)^{-1}$ followed by row normalization, yielding $Q^{(v)}=[q_{ij}^{(v)}]$.
We align $\{Q^{(v)}\}_{v=1}^{V}$ to the shared targets $P$ via
\begin{equation}
\mathcal{L}_{\mathrm{spec}}=\sum_{v=1}^{V}\mathrm{KL}\big(P\,\|\,Q^{(v)}\big),
\label{eq:spec-self-short}
\end{equation}
which turns the phase-consistent magnetic spectral signal into structured supervision and couples multi-view encoders with the curvature-refined shared geometry.

\subsection{Magnetic Laplacian Regularization and Label Consistency Learning}
\label{subsec:geometry-regularization}


\begin{algorithm}[t]
\small
\caption{Phase-Consistent Magnetic Spectral Learning for MVC}
\label{alg:cahmvc}
\begin{algorithmic}[1]
\REQUIRE Multi-view data $\{X^{(v)}\}_{v=1}^V$, \#clusters $K$
\ENSURE Cluster labels $y \in \{1,\dots,K\}^n$
\STATE Initialize $\{f^{(v)}_{\phi}, g^{(v)}_{\psi}\}_{v=1}^V$ and clustering heads producing $\{Q^{(v)}\}_{v=1}^{V}$.
\STATE Pretrain autoencoders by minimizing $\mathcal{L}_{\mathrm{rec}}$ to obtain $\{Z^{(v)}\}$.
\STATE For each view $v$: initialize anchors $A^{(v)}$ by $k$-means on $Z^{(v)}$, solve \eqref{eq:anchor-qp} to get $C^{(v)}$.
\STATE Construct anchor hypergraph $(H,w^{(0)})$ from $\{C^{(v)}\}$ and refine weights to $w^{(T_{\mathrm{RF}})}$ by \eqref{eq:ricci-update-short}; obtain $S'$ by \eqref{eq:anchor-affinity-short}.
\STATE Estimate top-anchor confidences from $\{C^{(v)}\}$, construct the antisymmetric flow $\tilde{F}$ and phase $\Theta$, then compute $(\tilde{A},L_{\mathrm{mag}})$, the magnetic embedding $U$, and shared targets $P$.
\STATE Lift $\tilde{A}$ to samples and construct $L_{\mathrm{cos}}$ according to Eqs.~\eqref{eq:sample-adj-complex-short}--\eqref{eq:lcos-short}.
\STATE \textbf{Stage I: geometry + spectral supervision}
\STATE Keep the shared geometry and targets fixed during both stages.
\FOR{epoch $=1$ to $T_{\mathrm{I}}$}
    \STATE Compute $\{Q^{(v)}\}$; update networks by minimizing
    $\mathcal{L}_{\mathrm{rec}} + \lambda_1 \mathcal{L}_{\mathrm{geom}} + \lambda_2 \mathcal{L}_{\mathrm{spec}}$.
\ENDFOR
\STATE \textbf{Stage II: label contrastive consistency}
\FOR{epoch $=1$ to $T_{\mathrm{II}}$}
    \STATE Compute $\{Q^{(v)}\}$; update networks by minimizing
    $\mathcal{L}_{\mathrm{rec}} + \mathcal{L}_{\mathrm{con}}$.
\ENDFOR
\STATE Recompute $Z,A,C,H,S',\Theta,U$ with the trained encoders and run $k$-means on $U$ to obtain $y$.
\end{algorithmic}
\end{algorithm}

\subsubsection{Magnetic-cosine Laplacian smoothing}
Let $\tilde{A} \in \mathbb{C}^{m \times m}$ be the curvature-refined magnetic adjacency in Eq.~\eqref{eq:mag-adjacency-short}, and let $H \in \mathbb{R}^{m \times n}$ be the anchor hypergraph incidence matrix.
We build a normalized sample-to-anchor association $M \in \mathbb{R}^{n \times m}$ from $H$ and lift $\tilde{A}$ to the sample level:
\begin{equation}
\delta=H^\top\mathbf{1}_m,\quad
M=\mathrm{diag}(\delta_\epsilon)^{-1}H^\top,\quad
\tilde{A}^{\mathrm{s}}=M\tilde{A}M^\top.
\label{eq:sample-adj-complex-short}
\end{equation}
We then form a real-valued magnetic-cosine affinity by combining magnitude and phase:

\begin{equation}
S_{\mathrm{mc}}=[|\tilde A^{\mathrm{s}}|\odot\cos(\arg\tilde A^{\mathrm{s}})]_+,
\qquad D_{\mathrm{s}}=\mathrm{diag}(S_{\mathrm{mc}}\mathbf1)+\epsilon I.
\end{equation}

where $[\cdot]_+$ denotes element-wise truncation to nonnegative values.
The corresponding normalized Laplacian is
\begin{equation}
L_{\mathrm{cos}} = I - D_{\mathrm{s}}^{-1/2} S_{\mathrm{mc}} D_{\mathrm{s}}^{-1/2}.
\label{eq:lcos-short}
\end{equation}

\paragraph{Geometry-aware label regularization.}
For each view $v$, let $Q^{(v)} \in \mathbb{R}^{n \times K}$ denote its soft cluster assignments.
We regularize $Q^{(v)}$ on the induced sample graph via the Laplacian quadratic form
\begin{equation}
\Omega\!\big(Q^{(v)}\big)
=
\frac{1}{nK}\,
\mathrm{tr}\!\big( Q^{(v)\top} L_{\mathrm{cos}} Q^{(v)} \big).
\label{eq:smooth-pen-short}
\end{equation}
Together with the anchor-based self-reconstruction and coefficient regularization, we define
\begin{equation}
\mathcal{L}^{(v)}_{\mathrm{geom}}
=
\big\| Z^{(v)} - A^{(v)} C^{(v)} \big\|_F^2
+ \gamma \big\| C^{(v)} \big\|_F^2
+ \lambda_c\,\Omega\!\big(Q^{(v)}\big),
\label{eq:geom-view-short}
\end{equation}
and aggregate it over views:
\begin{equation}
\mathcal{L}_{\mathrm{geom}} =
\sum_{v=1}^{V} \mathcal{L}^{(v)}_{\mathrm{geom}}.
\label{eq:geom-total-short}
\end{equation}

\subsubsection{Label contrastive consistency}

\paragraph{Cross-view alignment of cluster profiles.}
Since Stage~I aligns all $Q^{(v)}$ to the same $P$, cluster indices are synchronized before label-space contrastive refinement.
Let $q^{(v)}_j \in \mathbb{R}^{n}$ be the $j$-th column of $Q^{(v)}$.
We measure the similarity between two cluster profiles using cosine similarity:
\begin{equation}
\mathrm{sim}\big(q^{(v_1)}_j, q^{(v_2)}_k\big) =
\frac{\left\langle q^{(v_1)}_j, q^{(v_2)}_k \right\rangle}
{\big\| q^{(v_1)}_j \big\|_2 \big\| q^{(v_2)}_k \big\|_2}.
\label{eq:label-sim}
\end{equation}
For each ordered view pair $(v_1,v_2)$ with $v_1\neq v_2$ and each cluster index $j$, we treat
$\big(q^{(v_1)}_j, q^{(v_2)}_j\big)$ as a matched pair and
$\big(q^{(v_1)}_j, q^{(v_2)}_k\big)$ (for $k \neq j$) as mismatched, leading to
\begin{equation}
\mathcal{L}^{v_1,v_2}_{\mathrm{con}} =
- \frac{1}{K} \sum_{j=1}^{K}
\log
\frac{
\exp\big( \mathrm{sim}(q^{(v_1)}_j, q^{(v_2)}_j) / \tau_{\mathrm{con}} \big)
}{
\sum_{k=1}^{K}
\exp\big( \mathrm{sim}(q^{(v_1)}_j, q^{(v_2)}_k) / \tau_{\mathrm{con}} \big)
},
\label{eq:label-contrastive-mn-short}
\end{equation}
where $\tau_{\mathrm{con}}>0$ is the temperature.
Aggregating over all ordered view pairs gives
\begin{equation}
\mathcal{L}_{\mathrm{con}} =
\frac{1}{V(V-1)}
\sum_{\substack{v_1,v_2=1 \\ v_1 \neq v_2}}^{V}
\mathcal{L}^{v_1,v_2}_{\mathrm{con}}.
\end{equation}
We optimize the model in a two-stage scheme; the full procedure is summarized in Algorithm~\ref{alg:cahmvc}.

\begin{table}[t]
\centering
\small
\setlength{\tabcolsep}{3pt}
\renewcommand{\arraystretch}{1.05}
\begin{tabular}{@{}lccc@{}}
\toprule
Dataset & $V$ & $n$ & $K$ \\
\midrule
100Leaves (Cope et al.~\citeyear{cope2013leaves})
& 3 & 1600 & 100 \\

Caltech-5V (Xu et al.~\citeyear{xu2022mflvc})
& 5 & 1400 & 7 \\

Digit-Product (Cui et al.~\citeyear{Cui2024DCMVC})
& 2 & 30000 & 10 \\

ALOI (Geusebroek et al.~\citeyear{geusebroek2005aloi})
& 4 & 10800 & 100 \\

BDGP (Cai et al.~\citeyear{cai2012bdgp})
& 2 & 2500 & 5 \\

Fashion-MV (Xiao et al.~\citeyear{xiao2017fashionmnist})
& 3 & 10000 & 10 \\

UCI-3V (Duin~\citeyear{duin1998multiplefeatures})
& 3 & 2000 & 10 \\

Handwritten (Duin~\citeyear{duin1998multiplefeatures})
& 6 & 2000 & 10 \\

CUB (Wah et al.~\citeyear{wah2011cub})
& 2 & 600 & 10 \\

Multi-COIL-20 (Nene et al.~\citeyear{nene1996coil20})
& 3 & 1440 & 20 \\
\bottomrule
\end{tabular}
\caption{
Dataset statistics. $V$, $n$, and $K$ denote the numbers of views,
samples, and clusters, respectively.
}
\label{tab:dataset-stat}
\end{table}

\begin{table*}[!t]
\centering
\small
\setlength{\tabcolsep}{3.8pt}
\renewcommand{\arraystretch}{0.95}
\begin{tabular}{l*{5}{ccc}}
\toprule
Method
& \multicolumn{3}{c}{Caltech-5V}
& \multicolumn{3}{c}{CUB}
& \multicolumn{3}{c}{UCI-3V}
& \multicolumn{3}{c}{Fashion-MV}
& \multicolumn{3}{c}{Multi-COIL-20} \\
\cmidrule(lr){2-4}
\cmidrule(lr){5-7}
\cmidrule(lr){8-10}
\cmidrule(lr){11-13}
\cmidrule(lr){14-16}
& ACC & NMI & ARI
& ACC & NMI & ARI
& ACC & NMI & ARI
& ACC & NMI & ARI
& ACC & NMI & ARI \\
\midrule
Kmeans
& 78.51 & 68.31 & 61.71
& 75.83 & 75.01 & 61.15
& 90.05 & 83.66 & 80.71
& 77.84 & 75.47 & 68.03
& 65.48 & 83.81 & 63.26 \\

DEC’16
& \underline{88.64} & \underline{81.12} & 81.16
& 72.67 & 69.13 & 53.81
& 86.65 & 84.83 & 80.06
& 77.85 & 81.63 & 71.11
& 75.01 & 85.31 & 70.45 \\

DMCAG’23
& 83.64 & 75.41 & 66.95
& 67.42 & 63.64 & 50.16
& 85.85 & 82.26 & 76.09
& 84.01 & 90.12 & 82.13
& \underline{99.71} & \underline{99.68} & \underline{99.16} \\

AEMVC’25
& 75.21 & 66.43 & 58.32
& \underline{77.46} & \underline{76.51} & \underline{62.72}
& 93.23 & 91.42 & 89.71
& 67.81 & 55.82 & 49.21
& 91.53 & 94.15 & 91.31 \\

hubREP’25
& 68.79 & 61.31 & 54.37
& 67.01 & 72.45 & 56.78
& 83.81 & 86.56 & 80.02
& 73.12 & 73.61 & 62.16
& 68.33 & 80.32 & 65.34 \\

ALPC’25
& 86.29 & 74.82 & \textbf{84.29}
& 69.83 & 65.52 & 49.94
& 93.55 & 87.12 & 85.79
& 78.34 & 75.81 & 68.75
& 81.24 & 88.78 & 77.36 \\

STCMC-UR’25
& 69.34 & 66.84 & 45.37
& 76.31 & 71.47 & 56.73
& 51.24 & 59.79 & 36.81
& \underline{92.32} & \underline{93.21} & \underline{87.46}
& 81.76 & 91.56 & 77.31 \\

BONE’26
& 78.41 & 71.40 & 65.22
& 74.73 & 71.04 & 59.63
& 79.95 & 80.88 & 73.39
& 82.65 & 84.07 & 76.20
& \textbf{99.90} & \textbf{99.84} & \textbf{99.80} \\

DMVCS’26
& 85.51 & 80.78 & 76.27
& 67.90 & 70.90 & 55.84
& \textbf{97.21} & \textbf{93.64} & \textbf{93.89}
& 79.59 & 79.36 & 71.62
& 88.21 & 93.83 & 85.77 \\

Ours
& \textbf{90.23} & \textbf{82.45} & \underline{83.45}
& \textbf{85.12} & \textbf{81.44} & \textbf{73.45}
& \underline{96.25} & \underline{93.45} & \underline{93.65}
& \textbf{97.78} & \textbf{94.89} & \textbf{95.21}
& 99.32 & 98.45 & 97.23 \\

\midrule
& \multicolumn{3}{c}{BDGP}
& \multicolumn{3}{c}{Digit-Product}
& \multicolumn{3}{c}{Handwritten}
& \multicolumn{3}{c}{ALOI}
& \multicolumn{3}{c}{100Leaves} \\
\cmidrule(lr){2-4}
\cmidrule(lr){5-7}
\cmidrule(lr){8-10}
\cmidrule(lr){11-13}
\cmidrule(lr){14-16}
& ACC & NMI & ARI
& ACC & NMI & ARI
& ACC & NMI & ARI
& ACC & NMI & ARI
& ACC & NMI & ARI \\
\midrule
Kmeans
& 46.96 & 28.69 & 23.22
& 66.15 & 65.26 & 52.16
& 78.61 & 80.49 & 72.32
& 55.64 & 78.52 & 37.65
& 85.12 & 94.31 & 80.34 \\

DEC’16
& 93.73 & 86.31 & 86.81
& 74.61 & 75.56 & 63.07
& 93.81 & 87.57 & 86.81
& 53.81 & 86.45 & 39.93
& 82.94 & 92.12 & 78.25 \\

DMCAG’23
& 96.71 & 92.72 & 92.25
& \underline{98.31} & \underline{95.36} & 95.82
& 97.82 & \underline{95.16} & \underline{95.27}
& 82.94 & 88.52 & 51.31
& 89.36 & 94.87 & 69.52 \\

AEMVC’25
& 46.43 & 23.21 & 17.32
& 98.16 & 94.91 & \underline{95.97}
& 96.95 & 93.68 & 93.45
& \underline{90.23} & \underline{93.15} & \underline{84.34}
& \underline{91.43} & 96.23 & \underline{89.31} \\

hubREP’25
& \underline{98.11} & \textbf{95.12} & \underline{95.23}
& -- & -- & --
& 94.31 & 88.46 & 87.87
& 46.31 & 68.12 & 34.21
& 54.21 & 78.33 & 42.41 \\

ALPC’25
& 88.92 & 74.64 & 33.27
& 69.15 & 60.68 & 68.31
& 94.91 & 89.41 & 89.19
& 79.79 & 88.94 & 78.42
& -- & -- & -- \\

STCMC-UR’25
& 74.67 & 77.61 & 68.53
& 94.21 & 88.58 & 88.51
& 87.42 & 88.65 & 83.31
& 21.56 & 49.32 & 1.53
& 32.42 & 63.56 & 21.38 \\

BONE’26
& 97.60 & 94.17 & 94.64
& 95.63 & 89.57 & 90.64
& 92.94 & 89.13 & 87.36
& 66.69 & 83.05 & 57.50
& 72.65 & 89.00 & 66.92 \\

DMVCS’26
& 51.82 & 38.09 & 26.71
& 96.51 & 92.61 & 92.50
& \underline{97.84} & 94.95 & 95.25
& 85.64 & 92.77 & 81.64
& 90.96 & \underline{96.88} & 88.33 \\

Ours
& \textbf{98.81} & \underline{94.35} & \textbf{95.67}
& \textbf{98.89} & \textbf{96.78} & \textbf{96.32}
& \textbf{98.35} & \textbf{95.52} & \textbf{95.36}
& \textbf{91.36} & \textbf{93.32} & \textbf{84.56}
& \textbf{93.21} & \textbf{97.24} & \textbf{92.31} \\

\bottomrule
\end{tabular}
\caption{
\textbf{Overall clustering performance (\%).}
Rows correspond to methods, and column groups correspond to datasets.
Best results are shown in \textbf{bold}, and second-best results are underlined.
The symbol ``--'' indicates that the method failed to yield stable results
on the corresponding dataset.
}
\label{tab:overall_combined}
\end{table*}

\section{Experiments}
\label{sec:experiments}

This section presents overall comparisons with representative classical, deep, and optimization-guided
multi-view clustering baselines on diverse benchmarks, followed by component analyses and complementary studies to support the effectiveness of our phase-consistent magnetic spectral signal and the associated learning objectives.

\subsection{Experimental Setup}
\label{subsec:setup}

\paragraph{Datasets.}
We evaluate on ten widely-used multi-view clustering benchmarks summarized in Table~\ref{tab:dataset-stat}, covering both object/scene recognition and handwritten/attribute-style recognition tasks.
Specifically, 100Leaves and ALOI contain fine-grained object categories with multi-view feature descriptions; Caltech-5V, CUB, and Multi-COIL-20 are multi-view visual recognition benchmarks derived from image datasets; Digit-Product, Fashion-MV, UCI-3V, Handwritten, and BDGP provide complementary settings with different view numbers and cluster granularity.
Following common practice, each view is independently normalized by min--max scaling before training.

\paragraph{Baselines.}
We compare with nine representative baselines, including
K-means~\cite{kmeans} as a classical clustering method,
DEC~\cite{xie2016dec} as a representative deep single-view clustering method,
and recent multi-view clustering methods
DMCAG~\cite{dmcag23},
AEMVC~\cite{aemvc25},
ALPC~\cite{alpc},
hubREP~\cite{Xu_2025_CVPR},
STCMC-UR~\cite{hu2025stcmcur},
BONE~\cite{xu2026bone},
and DMVCS~\cite{li2026dmvcs}.
For each baseline, we use the official implementation
and follow the recommended settings under the same evaluation protocol.

\paragraph{Implementation details.}
Our model is implemented in PyTorch and optimized with Adam (learning rate $5\times10^{-4}$).
All experiments are conducted on a workstation with two NVIDIA GeForce RTX 5090 GPUs (32\,GB each).
We set the latent dimension to $d{=}10$ and follow the two-stage training procedure in Algorithm~\ref{alg:cahmvc}:
autoencoder pretraining (1000 epochs), Stage~I optimization with $\mathcal{L}_{\mathrm{rec}}+\lambda_1\mathcal{L}_{\mathrm{geom}}+\lambda_2\mathcal{L}_{\mathrm{spec}}$ (1000 epochs), and Stage~II optimization with $\mathcal{L}_{\mathrm{rec}}+\mathcal{L}_{\mathrm{con}}$ (100 epochs).
For anchor-hypergraph construction, we keep top-$r$ assignments per sample with $r{=}3$.
We set $T_{\mathrm{RF}}{=}3$, $\tau_{\mathrm{RF}}{=}0.1$, $q{=}0.25$, $\lambda_1{=}0.1$, $\lambda_2{=}1$, $\lambda_c{=}0.01$, and $\alpha{=}\tau_{\mathrm{con}}{=}1$; $m_a$ and $\gamma$ follow the dataset configuration.
Unless otherwise specified, we report the mean performance over 10 random seeds for all methods, including single-view baselines.

\begin{figure*}[t]
    \centering
    \begin{subfigure}[t]{0.32\textwidth}
        \centering
        \includegraphics[width=0.48\linewidth,trim=40mm 60mm 40mm 100mm,clip]{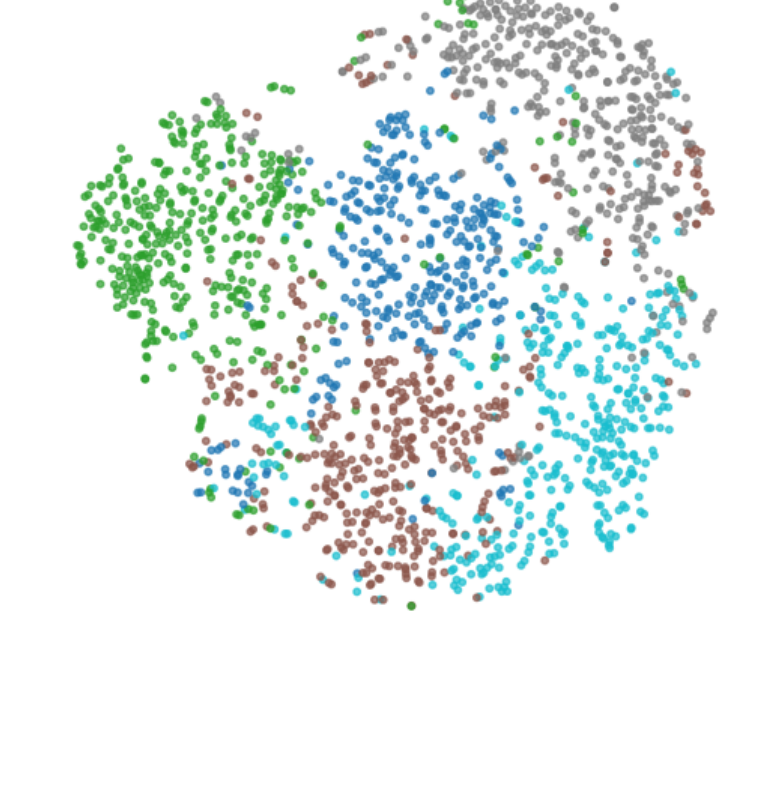}\hfill
        \includegraphics[width=0.48\linewidth,trim=40mm 60mm 40mm 100mm,clip]{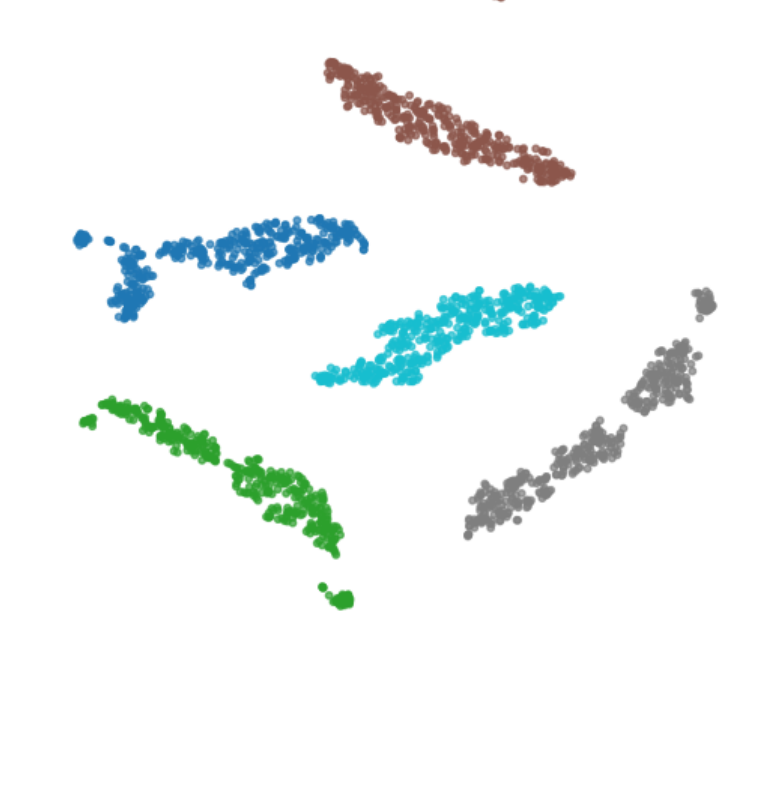}
        \caption{\texttt{BDGP}}
    \end{subfigure}\hfill
    \begin{subfigure}[t]{0.32\textwidth}
        \centering
        \includegraphics[width=0.48\linewidth,trim=40mm 60mm 40mm 100mm,clip]{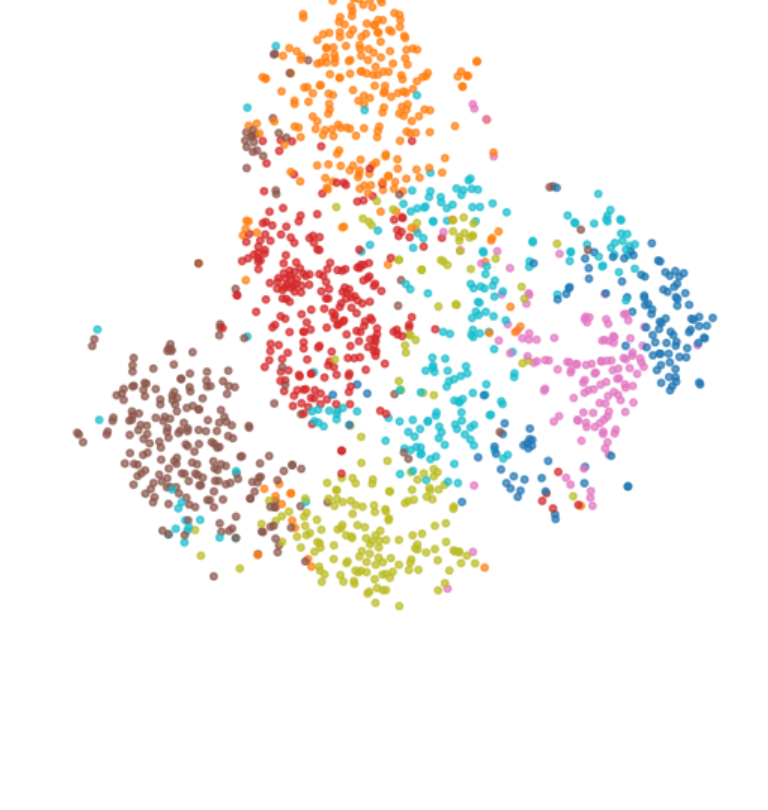}\hfill
        \includegraphics[width=0.48\linewidth,trim=40mm 60mm 40mm 100mm,clip]{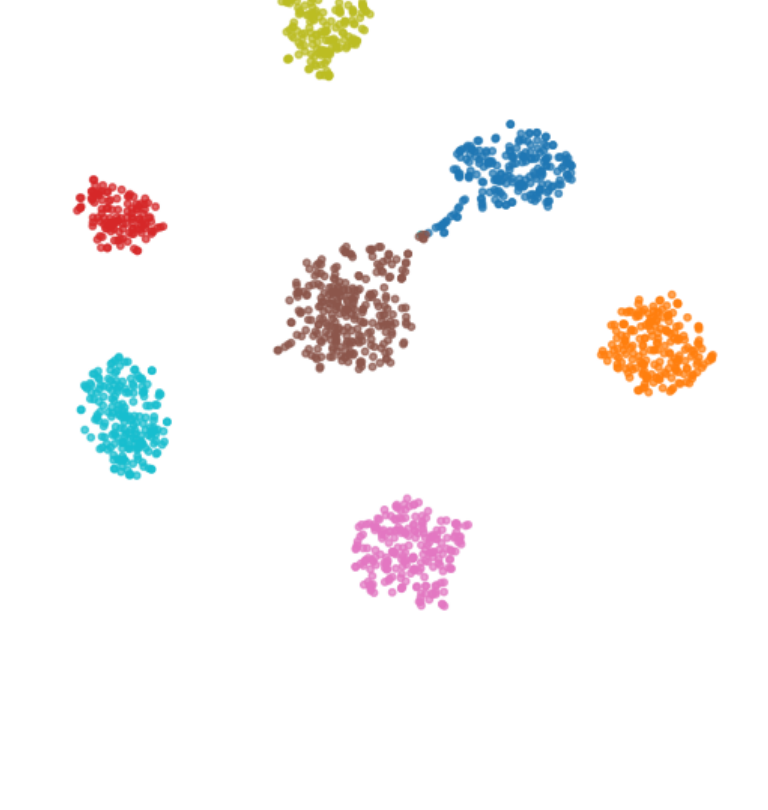}
        \caption{\texttt{Caltech-5V}}
    \end{subfigure}\hfill
    \begin{subfigure}[t]{0.32\textwidth}
        \centering
        \includegraphics[width=0.48\linewidth,trim=40mm 60mm 40mm 100mm,clip]{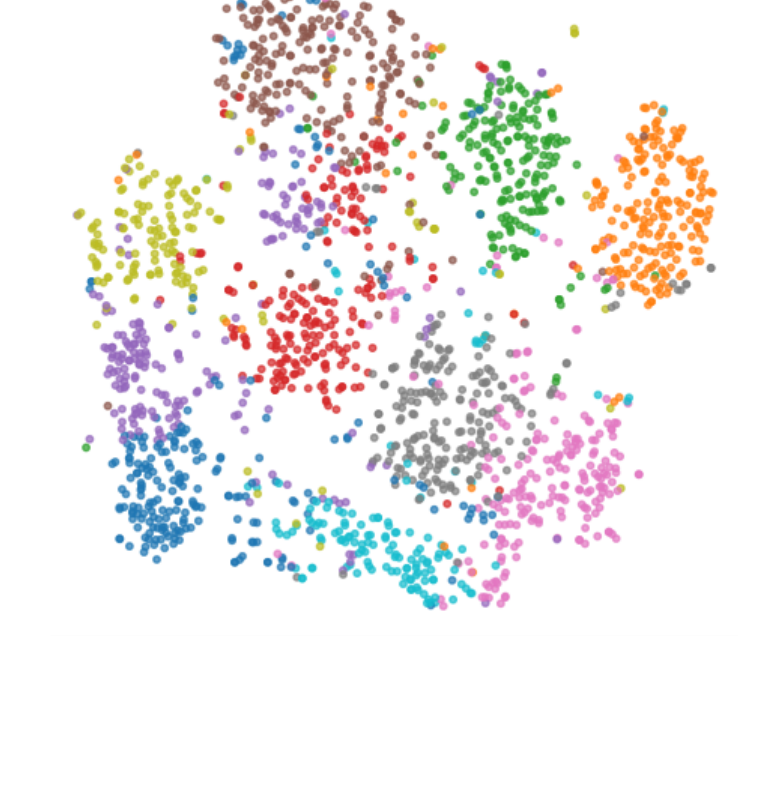}\hfill
        \includegraphics[width=0.48\linewidth,trim=40mm 60mm 40mm 100mm,clip]{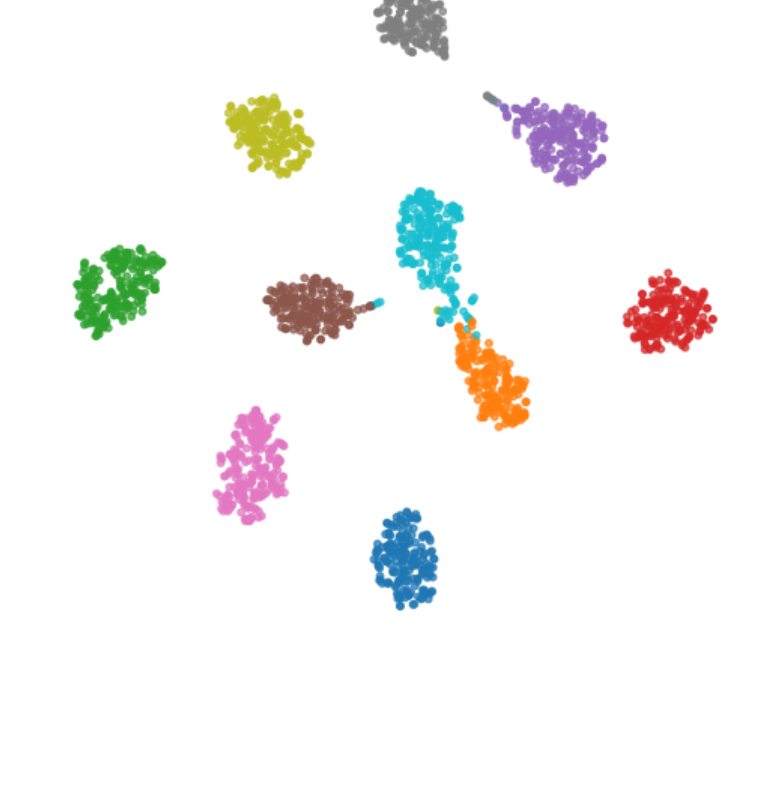}
        \caption{\texttt{Digit-Product}}
    \end{subfigure}
    \caption{\textbf{t-SNE visualizations before vs.\ after training.}
    Left/right panels show embeddings before/after training, respectively.}
    \label{fig:tsne_before_after}
\end{figure*}

\begin{figure}[t]
    \centering
    \begin{subfigure}[t]{0.48\columnwidth}
        \centering
        \includegraphics[width=0.80\linewidth]{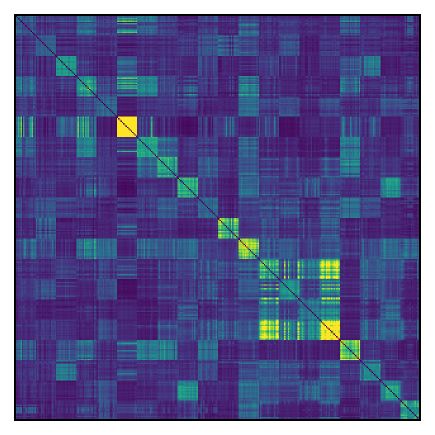}
        \caption{100Leaves}
    \end{subfigure}\hfill
    \begin{subfigure}[t]{0.48\columnwidth}
        \centering
        \includegraphics[width=0.80\linewidth]{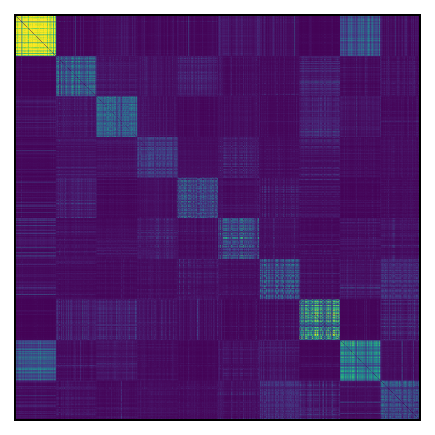}
        \caption{Handwritten}
    \end{subfigure}
    \caption{\textbf{Affinity heatmaps on two datasets.} Ground-truth labels are used only to reorder samples for visualization. Clearer diagonal blocks indicate stronger within-cluster connectivity, while suppressed off-diagonal entries indicate reduced cross-cluster leakage.}
    \label{fig:affinity_heatmaps}
\end{figure}

\subsection{Overall Performance}
\label{subsec:overall}

Table~\ref{tab:overall_combined} demonstrates that our method achieves
the best performance on 22 of the 30 dataset--metric combinations and
ranks second on five additional combinations.
It obtains the best results on all three metrics for CUB, Fashion-MV,
Digit-Product, Handwritten, ALOI, and 100Leaves, while also achieving
two best results on Caltech-5V and BDGP.
These results demonstrate broad robustness across datasets with
different scales, numbers of views, and cluster granularities.

The recent baselines also exhibit dataset-specific strengths.
DMVCS performs particularly well on UCI-3V, whereas BONE obtains the
best results on Multi-COIL-20.
Although our method is not uniformly dominant on every benchmark,
its leading performance on 22 metrics and top-two ranking on 27 out
of 30 metrics indicate a more consistent overall advantage across
diverse data distributions.
Fig.~\ref{fig:tsne_before_after} further shows that the learned
representations become more compact and better separated after
optimization, while Fig.~\ref{fig:affinity_heatmaps} presents clearer
block-diagonal structures with reduced cross-cluster leakage.

\subsection{Ablation: Magnetic Spectrum, Phase Structure, and Spectrum Stability}
\label{subsec:strong_ablation}

To isolate the contribution of the magnetic spectral mechanism,
we conduct a controlled phase ablation by fixing the shared magnitude backbone and varying only the phase construction and spectral operator.

We first construct the curvature-refined anchor geometry and obtain the shared nonnegative magnitude backbone $S'$.
On top of the same $S'$, we compare: (i) \emph{Real-Spec} by setting $\Theta=\mathbf{0}$ (equivalently $q{=}0$) and performing standard real-valued spectral embedding on $S'$;
(ii) \emph{Mag-Spec (ours)} by estimating an antisymmetric phase matrix $\Theta$ from cross-view assignments and extracting the embedding via the Hermitian magnetic Laplacian on
$\tilde{A}=S'\odot e^{i\Theta}$; (iii) \emph{Shuffled-Phase}, a control that keeps $S'$ unchanged but randomly permutes the per-sample anchor--confidence pairs across views before estimating $\Theta$, yielding $\Theta_{\text{shuf}}$; and (iv) \emph{Random-Phase}, which injects a random antisymmetric phase on the support of $S'$.

\paragraph{Spectrum stability metrics.}
We report stability indicators to explain \emph{why} the magnetic spectrum is more robust.
\emph{Eigengap} is defined as $\Delta_K=\lambda_{K+1}-\lambda_K$, where $\{\lambda_j\}$ are the eigenvalues (ascending) of the spectral operator; a larger $\Delta_K$ typically indicates clearer $K$-way separation \cite{vonluxburg2007spectral}.
Moreover, classical perturbation theory (e.g., Davis--Kahan) bounds the deviation of the top-$K$ invariant subspace by the perturbation magnitude scaled by $1/\Delta_K$, implying that larger $\Delta_K$ yields a more stable eigenspace \cite{davisKahan1970rotation,stewart1990matrix}.
To measure this stability directly, we compute a \emph{subspace distance} across seeds using principal angles:
for orthonormal bases $B^{(a)},B^{(b)}\in\mathbb{R}^{n\times 2K}$ obtained by thin QR factorization of $U$, the singular values of $(B^{(a)})^\top B^{(b)}$ satisfy $\sigma_j=\cos\theta_j$ \cite{bjorck1973angles},
and we report $\mathrm{Subspace}=\frac{1}{2K}\sum_{j=1}^{2K}\theta_j$ over all run pairs (smaller is better).

Table~\ref{tab:strong_phase_ablation} reports the controlled phase-ablation results on three datasets.
Across Fashion-MV, BDGP, and Caltech-5V, \emph{Mag-Spec} consistently outperforms \emph{Real-Spec} on ACC/NMI/ARI, and meanwhile yields a larger eigengap $\Delta_K$,
a smaller subspace distance (Sub), and a lower $k$-means inertia (Iner), indicating both clearer spectral separation and more stable/compact embeddings.
In contrast, \emph{Shuffled-Phase} and \emph{Random-Phase} markedly degrade all clustering metrics and stability indicators despite using the same magnitude backbone $S'$.
These controlled comparisons indicate that the gain is associated with \emph{structured, phase-consistent} estimation rather than arbitrary phase injection.

\begin{table}[t]
\centering
\small
\setlength{\tabcolsep}{1.95pt}
\renewcommand{\arraystretch}{1.05}
\begin{tabular}{lcccccc}
\toprule
Dataset / Variant
& ACC & NMI & ARI
& $\Delta_K\uparrow$
& Sub$\downarrow$
& Iner$\downarrow$ \\
\midrule

\multicolumn{7}{l}{\textbf{Fashion-MV}} \\
\quad Real-Spec ($q{=}0$)
& 95.91 & 92.72 & 93.14 & 0.089 & 0.46 & 1.72 \\
\quad \textbf{Mag-Spec (Ours)}
& \textbf{97.78} & \textbf{94.89} & \textbf{95.21}
& \textbf{0.125} & \textbf{0.31} & \textbf{1.34} \\
\quad Shuffled-Phase
& 92.83 & 88.12 & 88.91 & 0.061 & 0.62 & 2.15 \\
\quad Random-Phase
& 92.39 & 87.52 & 88.11 & 0.058 & 0.66 & 2.22 \\

\midrule
\multicolumn{7}{l}{\textbf{BDGP}} \\
\quad Real-Spec ($q{=}0$)
& 89.39 & 84.89 & 86.48 & 0.041 & 0.62 & 2.48 \\
\quad \textbf{Mag-Spec (Ours)}
& \textbf{98.81} & \textbf{94.35} & \textbf{95.67}
& \textbf{0.058} & \textbf{0.44} & \textbf{2.05} \\
\quad Shuffled-Phase
& 82.08 & 73.42 & 70.19 & 0.026 & 0.78 & 2.96 \\
\quad Random-Phase
& 81.62 & 72.83 & 69.61 & 0.024 & 0.81 & 3.02 \\

\midrule
\multicolumn{7}{l}{\textbf{Caltech-5V}} \\
\quad Real-Spec ($q{=}0$)
& 83.20 & 74.82 & 71.89 & 0.036 & 0.66 & 2.62 \\
\quad \textbf{Mag-Spec (Ours)}
& \textbf{90.23} & \textbf{82.45} & \textbf{83.45}
& \textbf{0.051} & \textbf{0.48} & \textbf{2.18} \\
\quad Shuffled-Phase
& 79.31 & 70.11 & 66.80 & 0.022 & 0.82 & 3.09 \\
\quad Random-Phase
& 78.94 & 69.65 & 66.21 & 0.021 & 0.85 & 3.15 \\

\bottomrule
\end{tabular}
\caption{
Controlled phase ablation under fixed $S'$ on three datasets.
$\Delta_K$ denotes the eigengap, Sub denotes the subspace distance,
and Iner denotes the $k$-means inertia.
}
\label{tab:strong_phase_ablation}
\end{table}

\subsection{Complexity and Efficiency}
\label{subsec:complexity}

Let $n$ be the number of samples, $V$ the number of views, and $m$ the number of anchors ($m \ll n$).
The sparse sample--anchor representation and anchor geometry scale with $O(nm)$ under a fixed sparsity level.
Most importantly, magnetic eigendecomposition is performed on the compact $m\times m$ anchor graph rather than an $n\times n$ sample graph, and the resulting spectral embedding is lifted to samples through the sample--anchor associations.
The sample-level regularizer in Eqs.~\eqref{eq:sample-adj-complex-short}--\eqref{eq:smooth-pen-short} may still involve an $n\times n$ affinity; hence, the principal computational saving comes from avoiding eigendecomposition on the sample graph rather than eliminating all quadratic sample-level operations.

\subsection{Parameter Sensitivity}
\label{subsec:parameter-sensitivity}

\paragraph{Loss-weight balancing.}
We analyze the sensitivity to the Stage-I loss weights $\lambda_1$ and
$\lambda_2$ for $\mathcal{L}_{\mathrm{geom}}$ and
$\mathcal{L}_{\mathrm{spec}}$, respectively, with the reconstruction
weight fixed to $1$ and Stage~II kept at the default setting.
We sweep $(\lambda_1,\lambda_2)$ on a log-scale grid and report results
averaged over 10 random seeds.

\begin{figure}[t]
    \centering
    \begin{subfigure}[t]{0.48\linewidth}
        \centering
        \includegraphics[width=\linewidth]{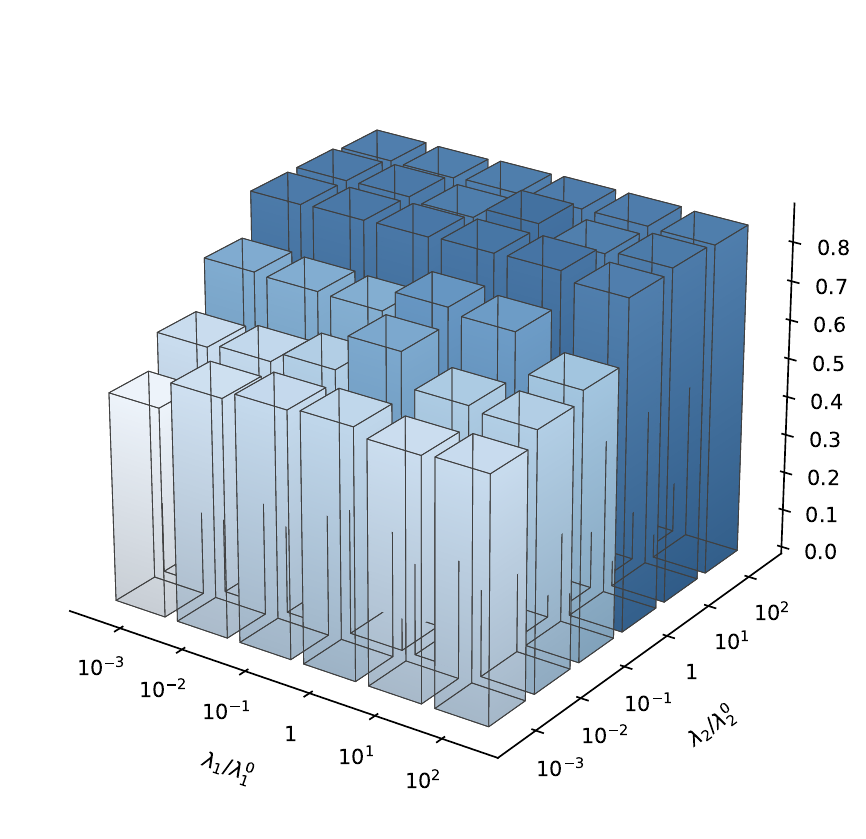}
        \caption{100Leaves}
    \end{subfigure}\hfill
    \begin{subfigure}[t]{0.48\linewidth}
        \centering
        \includegraphics[width=\linewidth]{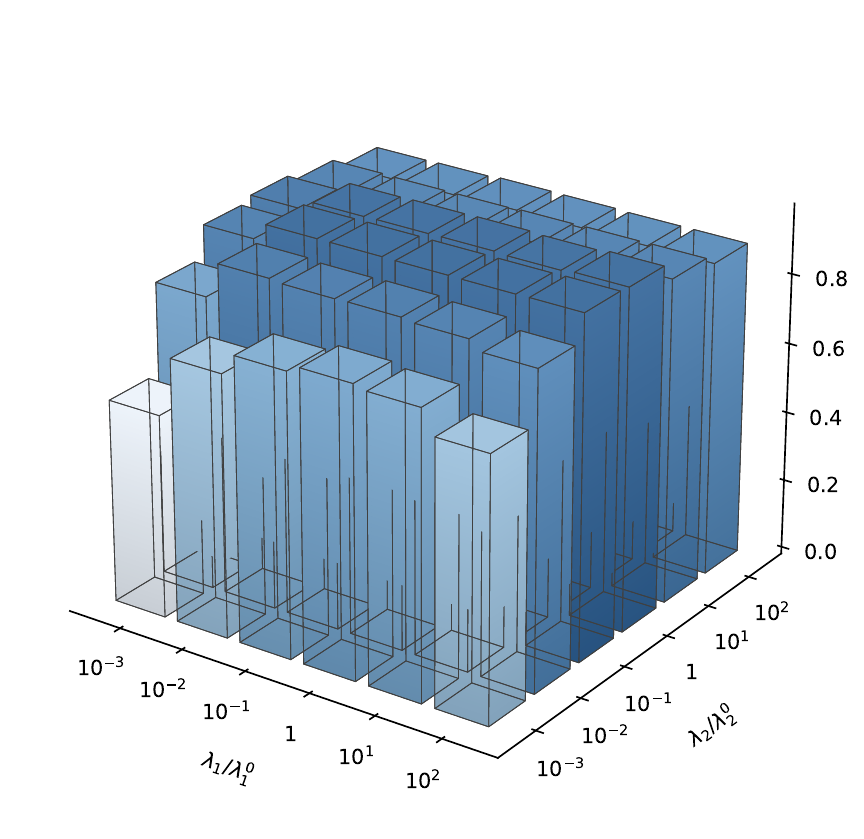}
        \caption{Fashion-MV}
    \end{subfigure}
    \caption{\textbf{Stage-I weight sensitivity.}
    ARI and ACC surfaces over $(\lambda_1,\lambda_2)$ on
    (a) 100Leaves and (b) Fashion-MV, respectively
    (axes in log scale).}
    \label{fig:lambda_sensitivity}
\end{figure}

As shown in Fig.~\ref{fig:lambda_sensitivity}, performance is more
sensitive to $\lambda_2$ and quickly stabilizes once $\lambda_2$ reaches
a reasonable range, while varying $\lambda_1$ causes relatively mild
changes. Overall, the broad high-performance plateau suggests that our
method is robust to loss-weight tuning.

\section{Conclusion}
\label{sec:conclusion}

Unsupervised MVC critically depends on a reliable shared structural signal under view discrepancy and noise. We propose \emph{Phase-Consistent Magnetic Spectral Learning} for MVC, which combines a nonnegative magnitude backbone with a phase term encoding confidence-directed cross-view flow, extracts a stable shared spectral signal via a Hermitian magnetic Laplacian, and uses it as structured self-supervision for robust learning. Experiments on ten benchmarks demonstrate leading or highly competitive performance against classical and recent MVC baselines, while extensive analyses confirm the effectiveness of the proposed phase-aware spectral modeling and its key components.

\clearpage
\bibliography{aaai27}

\end{document}